\documentclass{llncs}
\usepackage{graphicx}
\usepackage{amsmath}
\usepackage[caption=false,font=footnotesize]{subfig}
\usepackage{ctable}
\usepackage[strings]{underscore} 

\begin{document}
\title{Complex Network Construction for Interactive Image Segmentation using Particle Competition and Cooperation: A New Approach\thanks{Supported by the São Paulo Research Foundation – FAPESP (grant \#2016/05669-4).}}
\titlerunning{Complex Network Construction for Interactive Image Segmentation}
%
\author{Jefferson Antonio Ribeiro Passerini\orcidID{0000-0003-1382-9005} \and
Fabricio Breve\orcidID{0000-0002-1123-9784}}
\authorrunning{J. A. R. Passerini and F. Breve}
%
\institute{São Paulo State University, Rio Claro SP 13506-900, Brazil\\
\email{jefferson.passerini@gmail.com}, \email{fabricio.breve@unesp.br}}
\maketitle              
\begin{abstract}
In the interactive image segmentation task, the Particle Competition and Cooperation (PCC) model is fed with a complex network, which is built from the input image. In the network construction phase, a weight vector is needed to define the importance of each element in the feature set, which consists of color and location information of the corresponding pixels, thus demanding a specialist's intervention. The present paper proposes the elimination of the weight vector through modifications in the network construction phase. The proposed model and the reference model, without the use of a weight vector, were compared using 151 images extracted from the Grabcut dataset, the PASCAL VOC dataset and the Alpha matting dataset. Each model was applied 30 times to each image to obtain an error average. These simulations resulted in an error rate of only 0.49\% when classifying pixels with the proposed model while the reference model had an error rate of 3.14\%. The proposed method also presented less error variation in the diversity of the evaluated images, when compared to the reference model.
\keywords{Interactive Image Segmentation \and Machine Learning \and Semi-supervised Learning \and Particle Competition and Cooperation}
\end{abstract}
\section{Introduction}

Image segmentation is the process of identifying and separating relevant structures and objects from an image, for its later analysis and information extraction. This is considered one of the hardest tasks in image processing \cite{Gonzalez2010}.

The creation of image segmentation algorithms is considered an arduous task since it depends on the specific domain of the images or image groups. An algorithm designed for a specific domain usually has poor performance when applied to multiple images from different domains. Therefore, completely automated image segmentation still remains a challenge \cite{Gonzalez2010}.

Interactive image segmentation approaches rose as an alternative \cite{Boykov2001,Grady2006,Rother2004,Malmberg2011}, in which user input are used as clues to the algorithm. Many of these approaches are based on semi-supervised learning, as they are designed to handle scenarios with an abundance of unlabeled data and few labeled data, which are more onerous to be generated \cite{zhu2005Survey,Chapelle2006}.

The particle competition and cooperation (PCC) model \cite{Breve2012} is a graph-based supervised learning method. This model first converts the dataset into a non-weighted and non-orientated graph in which every data item corresponds to a node of the graph and its edges are generated from the similarity relations between the data items. The particles, which correspond to the labeled data, move in the graph cooperating with other particles of the same class and competing with particles of other classes.

Every group of particles aims to dominate the largest number of non-labeled nodes and to avoid the invasion of the labeled nodes by particles of other classes. At the end of the iterative phase, the boundaries are defined by the particles’ territories.

Many graph-based semi-supervised learning models are similar and share the same characteristics of operation, globally labeling the data \cite{zhu2005Survey}. The particle competition and cooperation model uses a local propagation of the labels approach, so its computational cost is close to linear ($O(n)$), while other approaches have cubic computation complexity ($O(n^3)$).

The PCC model has already been implemented in some important machine learning tasks such as the detection of overlapping communities \cite{Breve2013FuzzyCooperation}, learning with label noise \cite{Breve2015ParticleNoise}, active learning \cite{Breve2013ActiveLearning} and interactive image segmentation \cite{Breve2015ImagePcc,breve2015NonContiguous,breve2017BuildingNetworks}.

When applying the PCC model to image segmentation, a complex network is build based on the image to be segmented. Each pixel is represented as a node and the labeled pixels are also represented as particles \cite{Breve2015ImagePcc}. The edges are defined according to the similarity between each pair of pixels, measured by the Euclidean distance among features extracted from them, such as RGB and HSV components, and pixel localization. It is important to define a weight for each feature in this process according to their discriminative capacity in the image to be segmented. The weight vector directly affects the network construction and, therefore, it has a big impact on the PCC segmentation accuracy.

So far, the proposed methods for defining features weights work well in some images and fail in others, demonstrating the difficulty of this process. In \cite{breve2015AutoFeature}, the weights are calculated with automatic methods based on the average, standard deviation and histogram of the labeled pixels. Each method was able to increase the accuracy in some tested images but failed in others. In \cite{breve2017BuildingNetworks}, a network index is proposed to evaluate candidate networks. A genetic algorithm is used to optimize the network index, thus evolving a good network. However, the optimization process is time-consuming and the network index is based on assumptions that do not hold for all images.

This paper proposes the elimination of the weight vector. To achieve this, it changes the complex network construction in the following way: (a) a different set of features is used, (b) a new form of user annotation is introduced, (c) a new approach is used to define the edges among network nodes, and (d) the particle influence on the network is measured before the competition process starts.

Simulations were made using $151$ real-world images. $49$ of them are taken from the GrabCut dataset, $99$ from the PASCAL VOC dataset, and $3$ images from the Alpha matting dataset. These are the same images used in \cite{Gulshan2010}. The segmentation efficacy is calculated through the comparison between the segmented image using the proposed model, the reference model, and versions of the images segmented by experts and made available in the data sets.

The remaining of this paper is organized as follows: Section \ref{sec:ispcc} describes the particle competition and cooperation model applied to image segmentation. Section \ref{sec:newcomplexnetwork} describes the proposed method to build the complex networks. Section \ref{sec:Experiments} describes how the experiments were performed. In Section \ref{sec:resultsdiscussion} the results are presented and discussed. Finally, in Section \ref{sec:Conclusions} we draw some conclusions.

\section{Image segmentation using particle competition and cooperation}
\label{sec:ispcc}

Previous papers demonstrate the PCC model applied to image segmentation \cite{Breve2015ImagePcc,breve2015NonContiguous,breve2015AutoFeature,breve2017BuildingNetworks}. The image is first converted into a complex network where nodes represent image pixels and edges represent the similarity between them, measured through the Euclidean distance among weighted features extracted from those pixels. The weight vector defines the importance of each feature in the set, for each processed image. Unfortunately, a different set of weights is required for each image to be processed \cite{breve2015NonContiguous,breve2017BuildingNetworks}.

Particles are created for each pixel that was labeled by the human specialist. During the iterative phase, these particles walk in the network cooperating with other particles of the same class and competing against particles from other classes for the possession of the unlabeled nodes. They use a random-greedy rule to decide the next node to visit.

Each node has a set of domination levels. Each of these levels corresponds to a class of the image. When a particle visits a node, it can increment the dominance level of its class on the node, while reducing the domination level of the other classes. Every particle has a strength level which is altered according to the nodes it visits. Every particle also has a distance table that contains the distance, measured in hops, from every visited node to its origin node. This table is dynamically updated during the particle's walk through the network.

In \cite{breve2015AutoFeature,breve2017BuildingNetworks}, a set of 23 features is extracted from the pixels: (1-2) pixel location components (line, column), (3-5) RGB components, (6-8) HSV components, (9-14) the average of the RGB and HSV components on the pixel and its neighbors, (15-20) the standard deviation of the RGB and HSV components on the pixel and its neighbors, (21-23) the ExR, ExG, and ExB components.

For the features averages and standard deviation, the $8$-connected neighbors of the pixel are considered, except for the borders. All the extracted components are normalized to have a mean $0$ and a standard deviation of $1$. The HSV components were calculated as described by \cite{Smith1978ColorPairs}, and the components ExR, ExG, and ExB were obtained from the RGB components as described in \cite{Lichman2013}. Each feature can be emphasized or not through a weight vector $\lambda$ during the Euclidean distance calculation.

The network is represented by a non-directed and unweighted graph. Therefore, $G = \{V,E\}$, where $V=\{v_{1}, v_{2}, \ldots, v_{n}\}$ is the set of nodes and $E=\{v_{i},v_{j}\}$ determinates the set of edges. Each node $v_{i}$ corresponds to a pixel $x_{j}$. Two nodes $v_{i}$ and $v_{j}$ are connected if $v_{j}$ is among the k-nearest neighbors of $v_{i}$, considering the Euclidean distance features of $x_{i}$ and $x_{j}$, otherwise, they are disconnected.

For each node $v_{i} \in \{v_{1}, v_{2}, \ldots, v_{L}\}$ correspondent to a labeled pixel $x_{i} \in X_{L}$, a particle $p_{i}$ is generated and its initial position is defined as $v_{i}$. Each particle $p_{j}$ has a variable $p^{w}_{j}(t) \in \left[0,1\right]$ to store its strength, which determines the impact generated by the particle to the visited node. The initial strength is always set to its maximum value $p^{w}_{i}(0)=1$.

Each particle has a table of distances which is dynamically updated during the movement of a particle, in which the distance between the particle's initial position and the visited nodes are stored. It is determined by $p^{d}_{j}(t) = \{p^{d_{1}}_{j}(t), p^{d_{2}}_{j}(t), \ldots, p^{d_{n}}_{j}(t)\}$, where each element $p^{d}_{j}(t) \in \left[0,n-1\right]$ corresponds to the measured distance between the particle's original node $p_{j}$ and the node $v_{i}$.

Each node $v_{i}$ has a vector variable $v^{w}_{i}(t) = \{v^{w_{1}}_{i}(t), v^{w_{2}}_{i}(t), \ldots, v^{w_{n}}_{i}(t)\}$ where each element $v^{w}_{i}(t) \in \left[0,1\right]$ represents the domination levels of the team $l$ over the node. The sum of this vector is always a constant,
\begin{equation}\label{eq:relacao_maxima_valor_no}
	\sum^e_{l=1}v^{w_l}_i = 1 \textrm{.}
\end{equation}
This occurs because when a particle increases its dominance level on the node, the domination of the other groups decrease in the same proportion for all the classes, which is determined as law of conservation of forces.

The initial domination levels $v^{w}_{i}(t)$ of each node $v_{i}$ is configured as:
\begin{equation}\label{eq:vet_no_dominio}
	\begin{aligned}
		W_{i}^{wl} = \left\{\begin{matrix}
			1 & \textrm{ if } x_{i} \textrm{ is labeled and } y(x_{i}) = l \\
			0 & \textrm{ if } x_{i} \textrm{ is labeled and } y(x_{i}) \neq l \\
			\frac{1}{c} & \textrm{ if } x_{i} \textrm{ is not labeled }
		\end{matrix}\right. \textrm{.}
    \end{aligned}
\end{equation}

During the iterations, the domain values vary according to the equation:
\begin{equation}\label{eq:variacao_dominio}
	\begin{aligned}
		v_i^{w_{l}}\left(t+1\right) = \left\{\begin{matrix}
			max\left\{0, v_i^{w_{l}}\left(t\right) - \frac{\Delta_vp^w_j\left(t\right)}{c-1}\right\} & \textrm{if } x_{i} \textrm{ is unlabeled and }l \neq p^{f}_{j} \\
			v_i^{w_{l}}\left(t\right) + \sum_{q \neq c} v_i^{w_{q}}\left(t\right) - v_i^{w_q}\left(t+1\right) & \textrm{if } x_i \textrm{ is unlabeled and } l = p^{f}_{j} \\
			v_i^{w_{l}}\left(t\right) & \textrm{ if } x_{i} \textrm{ is labeled}
		\end{matrix}\right. \textrm{,}
    \end{aligned}
\end{equation}
where $p^{f}_{i}$ represents the team of the particle $p_j$. Each particle will change the node its visiting $v_{i}$, increasing the domination level of its class $(v^{w_{l}}_{i}(t), l=p^{f}_{j})$ on it, and decreasing the domination levels of other classes $(v^{w_{l}}_{i}(t), l \neq p^{f}_{j})$, always respecting the law of conservation of forces.

The particle's force depends on the domination level of its team on the visited node. Therefore, in every iteration, the force of the particle $p^{w}_{j}(t)$ is updated according to the equation: $p^{w}_{j}(t+1)=v^{w_{l}}_{i}(t+1)$, where $v_{i}$ is the target node and $l=p^{f}_{j}$, with $l$ being the label of the particle's team $p_{j}$, consequently each particle $p_{j}$ has its strength $p^{f}_{j}$ configured with the domination level value of its team $w^{w_{j}}_{i}$ in the node $v_{i}$.

The model conditions the particle's force to a reducer $\Delta_{v}$, in this way it is possible to control the speed in which the particles will dominate the unlabeled nodes.

When the node is visited, the distance table is updated according to the following equation:
\begin{equation}\label{eq:atuaz_tab_dist}
	\begin{aligned}
		p^{d_k}_j\left(t+1\right) = \left\{\begin{matrix}
			p^{d_i}_j\left(t\right) + 1 & \textrm{ if } p^{d_i}_j\left(t\right)+1 < p^{d_k}_j\left(t\right) \\
			p^{d_k}_j\left(t\right) & \textrm{otherwise}
		\end{matrix}\right. \textrm{.}
    \end{aligned}
\end{equation}
This way, when the particle moves from the current node to the target node, it checks the distance table and, if the distance from the target node is larger than the current node's distance plus 1, the table is updated. This way, it is not necessary to know the distances for every node a priori.

In every iteration, the particle must choose which node it will visit among the $k$-nearest neighbors from the current position, this movement is based in one of two rules: random rule or greedy rule.

In the random rule the particle randomly chooses, with equal possibilities, any of the neighboring vertices from the position in which the particle is located. This rule does not take into account the levels of domain or distances of the original node, being useful for exploration and acquisition of new nodes. It is defined by the equation:
\begin{equation}\label{eq:part_random}
	p\left(v_i|p_j\right) = \frac{W_{qi}}{\sum^n_{\mu=1} W_{q\mu}} \textrm{,}
\end{equation}
where $q$ is the node's index where the current particle $p_{j}$ is located, therefore $W_{qi} = 1$ if there is an edge between the current node and any other node $v_{i}$ or else $W_{qi} = 0$.

In the greedy rule, the particle randomly chooses any one of the neighboring nodes from the current node, with probabilities calculated directly proportional to the dominance level of this particle's team in every neighboring node, and inversely proportional to the distance of these neighbors to the particle's local node. This movement is useful for the team's territory defense. It is defined by the following equation:
\begin{equation}\label{eq:part_greedy}
	p\left(v_i|p_j\right) = \frac{W_{qi}v^{w_l}_{i}\left(1+p^{d_{i}}_j\right)^{-2}}
	               {\sum^n_{\mu=1}W_{q\mu}v^{\omega_l}_{\mu}\left(1+p^{d_{\mu}}_{j}\right)^{-2}} \textrm{,}
\end{equation}
where $q$ is the node's index where the current particle $p_{j}$ is located, and $l = p^{f}_{j}$, with $p_{j}^{f}$ being the class label of the particle $p_{j}$.

At each iteration, every particle has the probability $P_{grd}$ of choosing the greedy movement and the probability $1-P_{grd}$ of choosing the random movement, with $0 \leq p_{grd} \leq 1$.

The algorithm stops when it reaches the maximum number of iterations $(maxIte)$, or when the node's domination levels reach some stability. The stability is inferred when, on average, the nodes had improved below a threshold ($controlStop$) in their maximum domination level in a given iteration interval $(maxStop)$.

\section{Constructing a new complex network for the model}
\label{sec:newcomplexnetwork}

Based on what has been presented in previous papers, this paper proposes a change in how the complex network is built and the elimination of the weight vector that defines the importance of each feature within the set.

\begin{figure}[htb]
\centering
\setlength\tabcolsep{1.5pt}
\begin{tabular}{cc}
\subfloat{\includegraphics[height=4cm]{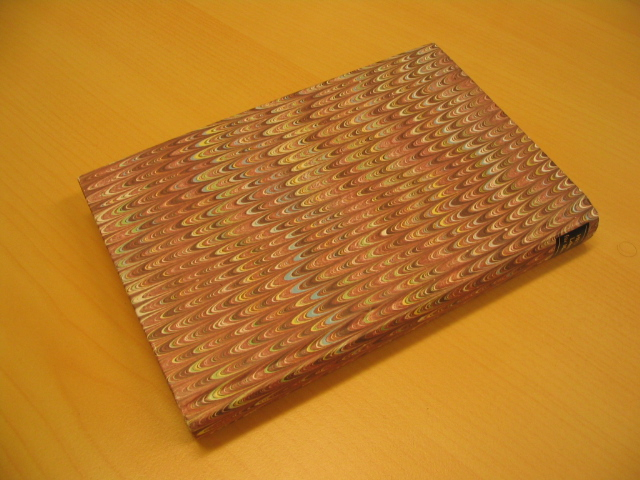}} &
\subfloat{\includegraphics[height=4cm]{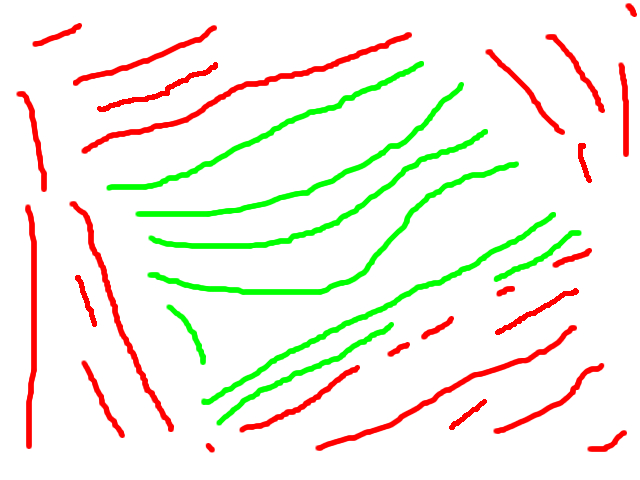}} \\
(a) & (b) \\
\subfloat{\includegraphics[height=4cm]{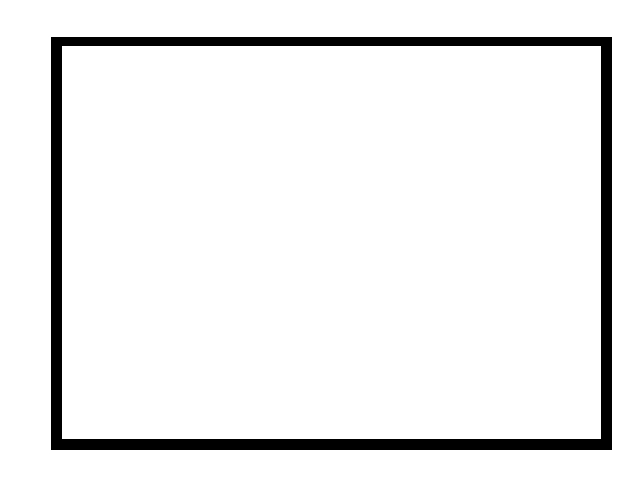}} &
\subfloat{\includegraphics[height=4cm]{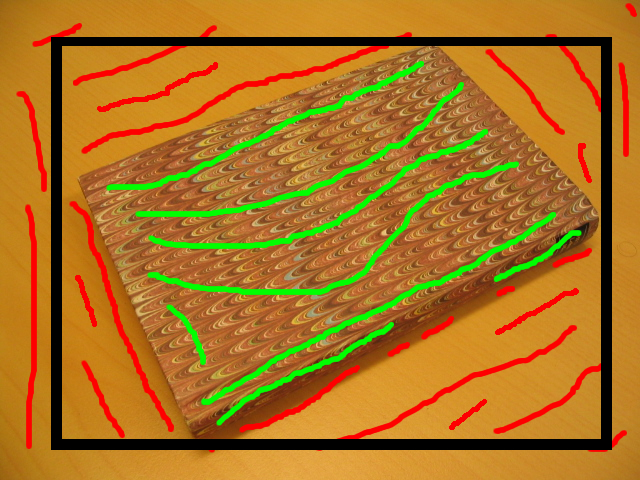}} \\
(c) & (d)
\end{tabular}
\caption{Input information: (a) the real-world images to be segmented, (b) the ``scribbles'' provided by the user, (c) the cut polygon, and (d) overlay image for visualization only.}
\label{fig:InputInformation}
\end{figure}

The original model receives the image to be segmented (for example: Figure \ref{fig:InputInformation}a) and the user input (Figure \ref{fig:InputInformation}b). In this new approach, it is also possible to delimit, in the image, the region of interest where the object to be segmented is found (Figure \ref{fig:InputInformation}c), to reduce the processing scope.

Figure \ref{fig:InputInformation}d displays a composition of the algorithm's input information. The cut polygon represented in Figure \ref{fig:InputInformation}c is optional to the model because there can be different situations that the cut in the original image is not of interest since the segmentation object of choice is set across most of the image. Regardless of the usage of the cut polygon by the algorithm or not, the complex network assembly process is the same.

The input image passes through a bicubic interpolation process to reduce its dimension (amount of pixels) allowing the processing of larger images. Later, the bilinear interpolation algorithm is used for information recomposition after the segmentation processing, as described in \cite{Pedrini2007,Prajapati2012,Getreuer2011}.

\begin{figure}[htb]
\centering
\setlength\tabcolsep{1.5pt}
\begin{tabular}{cccc}
\subfloat{\includegraphics[height=2.8cm]{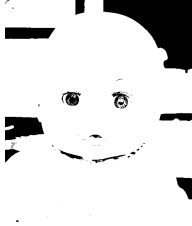}} &
\subfloat{\includegraphics[height=2.8cm]{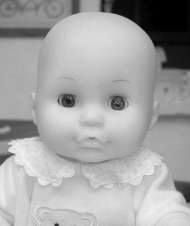}} &
\subfloat{\includegraphics[height=2.8cm]{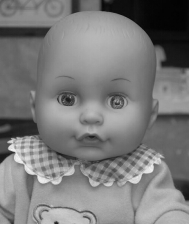}} &
\subfloat{\includegraphics[height=2.8cm]{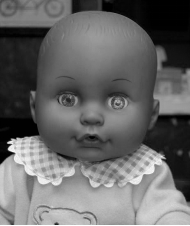}} \\
(a) & (b) & (c) & (d) \\
\subfloat{\includegraphics[height=2.8cm]{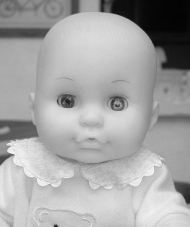}} &
\subfloat{\includegraphics[height=2.8cm]{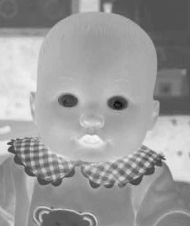}} &
\subfloat{\includegraphics[height=2.8cm]{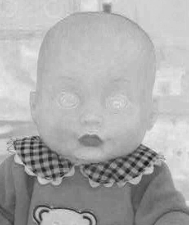}} &
\subfloat{\includegraphics[height=2.8cm]{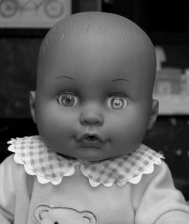}} \\
(e) & (f) & (g) & (h)
\end{tabular}
\caption{Proposed algorithm's extracted features representation: (a) the binarization made with Otsu’s algorithm, (b-d) the RGB components, respectively, (e) the V component from the HSV system, (f-h) the ExR, ExG, and ExB components, respectively.
}
\label{fig:DataExtracted}
\end{figure}

Once the image is reduced, the extraction of the image's pixels features set is started. This new proposed approach uses less features than its predecessors: (1-2) pixel location components (line, column), (3-5) RGB components, (6) only the V (value) component of the HSV system \cite{Smith1978ColorPairs}, (7-9) the color components ExR, ExG, ExB defined in \cite{Lichman2013}, and (10) a new feature extracted using Otsu’s binarization algorithm \cite{Otsu1979}. Figure \ref{fig:DataExtracted} shows graphic representations of each color feature extracted from an image from the Grabcut data set.

The features are normalized to have an average of $0$ and a standard deviation of $1$. Thereafter occurs the non-oriented graph generation, which will represent the pixels of the image through its nodes, while the edges represent the similarity relationships between the pixels. As has been previously stated, the similarity relationships are determined by measuring the Euclidean distance among the extracted features from the image to be segmented.

In the reference model, each node is connected to its $k$-nearest neighbors, considering the Euclidean distance among pixel features, and $k$ is a parameter set by the user. In the proposed model, $k$ is fixed, and each node is connected to its $192$ nearest neighbors. Another $8$ connections are made based in the pixel spatial neighborhood, defined by a $3 \times 3$ window, i.e., the corresponding node will be linked to the nodes corresponding to its $8$ physically adjacent pixels. This change was introduced to consider the pixel physical neighborhood when creating a complex network since the original approach cannot guarantee they will be connected.

Another alteration is that the proposed model verifies whether two linked nodes are of different classes and, if this happens, the new edge will not be generated. Nodes can only be linked with nodes of the same class or with unlabeled nodes.

Once the complex network is defined, this approach considers the influence of a previously labeled pixel (particle) in relation to its physical neighbors in the original image. This alteration also aims to reinforce the relationship information among the image's pixels before the iterative phase, but now in relation to the pixels labeled by the user in a $5 \times 5$ window.


It is considered that a particle (labeled pixel) influences other nodes in the network up to the distance of $2$ hops in the structure. It must be emphasized that, in this approach, the distance also takes into consideration the spatial distance of the pixel present in the image. This alteration aims to value the pixel's relationships of the spatial neighborhood in the image to be analyzed since a neighboring unlabeled pixel tends to be part of the same class.

\begin{figure}[htb]
\centering
\setlength\tabcolsep{1.5pt}
\begin{tabular}{c}
\subfloat{\includegraphics[height=5cm]{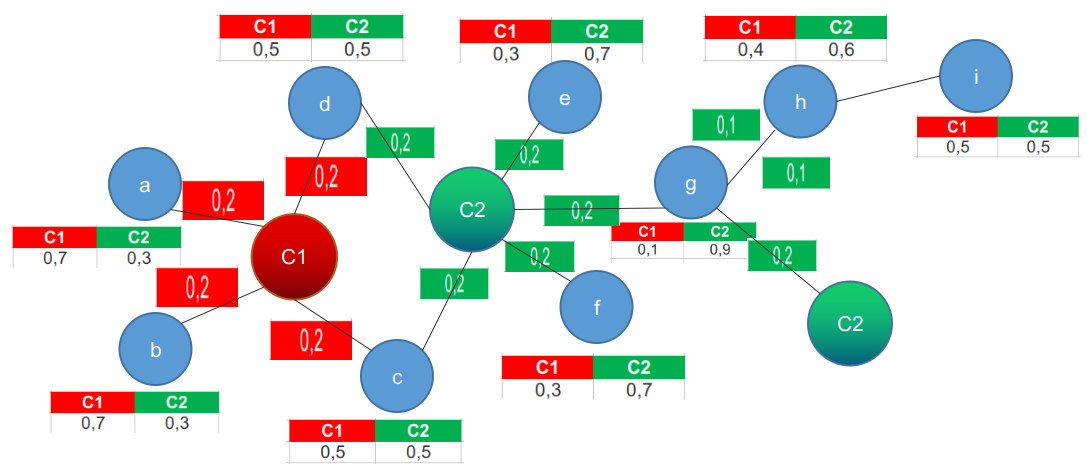}}
\end{tabular}
\caption{Representation of the labeled pixels (particles) influence on the spatial neighborhood of the original image.}
\label{fig:ParticleInfluence}
\end{figure}

Therefore, unlabeled nodes which are $1$ hop away from a labeled node will have an increment of $0.2$ in its domination vector, referent to the nearby particle's class. There is the possibility that the unlabeled node suffers the influence in its domination vector of more than one nearby particle. Unlabeled nodes which are $2$ hops away from a labeled node suffer less influence in their domination vector. The increment is only $0.1$ in the nearby particle's class. This process is illustrated in Figure \ref{fig:ParticleInfluence}.

This approach respects the law of conservation of forces, previously described. If a pixel has its domination vector completely dominated by a given team of particles, it will still be subject to changes during the iterative phase, differently from the labeled nodes, which have fixed domination levels.

After the network construction and initial settings with the proposed approach, the iterative phase takes place exactly as described in earlier particle competition and cooperation models.

\section{Experiments}
\label{sec:Experiments}

Two models were implemented to perform the tests with the proposed approach in this paper. The first one is the reference model \cite{Breve2015ImagePcc,breve2015NonContiguous,breve2015AutoFeature,breve2017BuildingNetworks}, with the $23$ features already described. The weight vector $\lambda$ was defined so all the features had the same weight. The reference model receives the interpolation approach as an increment in its execution to reduce the size of the images to be processed. The second is the proposed model, as described in the previous session.

$151$ real-world images taken from the GrabCut dataset, the PASCAL VOC dataset, and from the Alpha matting dataset are used to evaluate both models. These are the same images used in \cite{Gulshan2010}. The markings (labels) defined for the tests and the cut polygons used in this work are available at Github\footnote{Available at https://github.com/jeffersonarpasserini/dataset-interactive-algorithms.git}. In our simulations, each image was processed $30$ times by each model to calculate the models’ average performance. In total $4,530$ executions were performed with each algorithm.

The error rate is determined by comparing the algorithms' output with the ground-truth images. Notice that the ground-truth images have some pixels (in grey) representing divergence in classification among the specialists. Those pixels were not considered in the error calculation.

Both models were configured to use $200$ neighbors, $\Delta_{v}=0.1$, $P_{grd}=0.5$, $maxIte=1,000,000$, $maxStop=15,000$, and $controlStop=0.001$. The same images and annotated pixels were used for both models.

The simulations were performed in an Asus laptop model S46C configured with a Quadcore 2Ghz Intel i7 processor, model 3537U, with 16GB of RAM, running the Linux Deepin 15.5 Desktop operating system. The models' development was made using the Python version 2.7 programming language.

\section{Results and discussion}
\label{sec:resultsdiscussion}

The experiments were performed as discussed in Section \ref{sec:Experiments}. It was observed that in five images from the data set it was not possible to apply the cut polygon feature, due to the subject being segmented occupying most of the original image. Those images are presented in Table \ref{Tab:ResPolygonCut} with the error rates achieved by both models.

\begin{table}
\centering
\caption{Error rates in the five images that did not use the cut polygon resource, as achieved by the proposed model and the reference model.}
\begin{tabular}{lrr}
\hline
Image name       & Proposed & Reference \\
\hline
Baby\_2007\_006647 & 1.17\%                       & 4.57\%                         \\
cross              & 0.48\%                       & 1.79\%                         \\
gt02               & 0.52\%                       & 1.27\%                         \\
gt07               & 0.21\%                       & 0.64\%                         \\
gt13               & 1.08\%                       & 2.11\%                         \\
\hline
Average              & 0.64\%                       & 1.72\%                         \\
\hline
\end{tabular}
\label{Tab:ResPolygonCut}
\end{table}

\begin{figure}
\centering
\setlength\tabcolsep{1.5pt}
\begin{tabular}{cccc}
\multicolumn{4}{c}{baby_2007_006647} \\
\subfloat{\includegraphics[height=2cm]{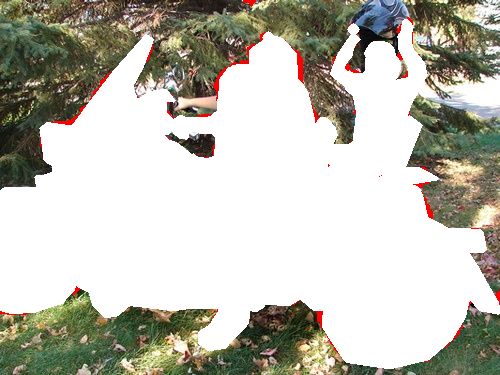}} &
\subfloat{\includegraphics[height=2cm]{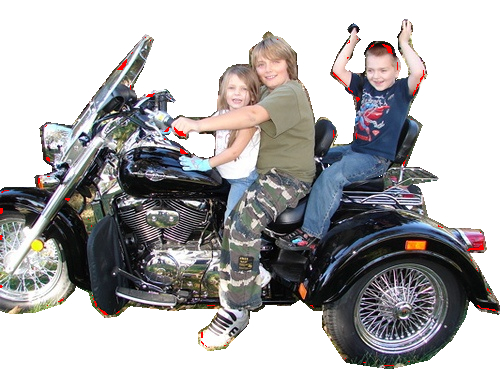}} &
\subfloat{\includegraphics[height=2cm]{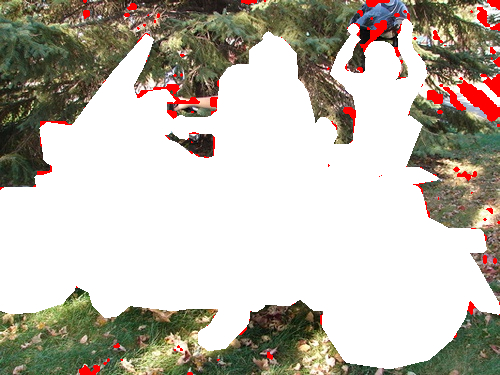}} &
\subfloat{\includegraphics[height=2cm]{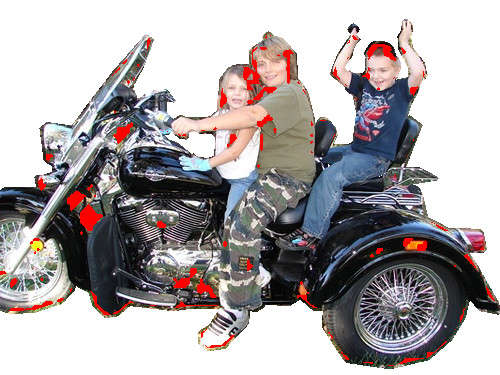}} \\
\multicolumn{2}{c}{Proposed Method - Error Rate: 1.17\%} &
\multicolumn{2}{c}{Reference Method - Error Rate: 4.57\%} \\
\\
\multicolumn{4}{c}{gt02} \\
\subfloat{\includegraphics[height=2cm]{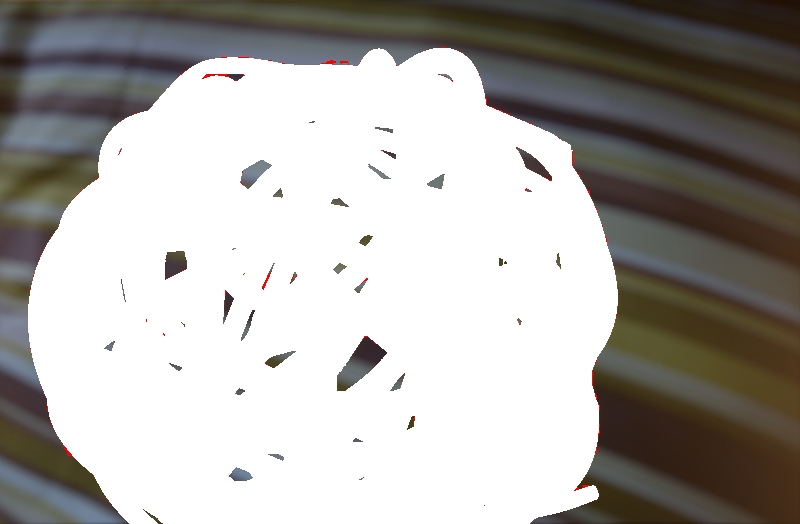}} &
\subfloat{\includegraphics[height=2cm]{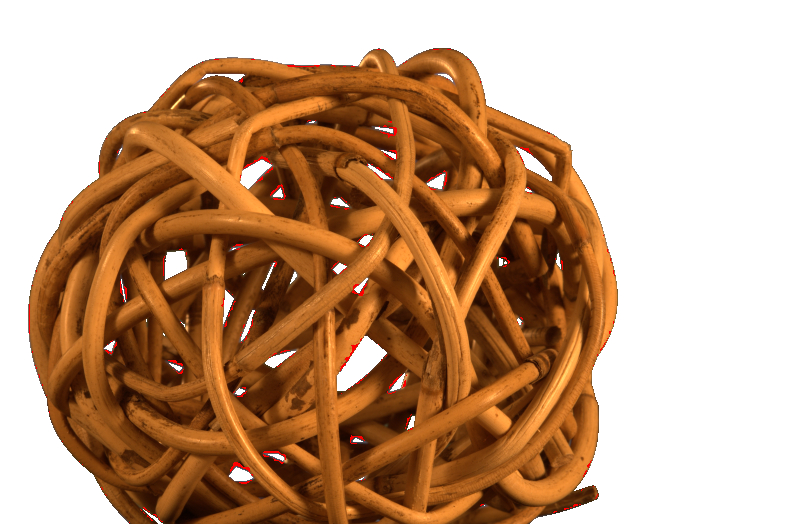}} &
\subfloat{\includegraphics[height=2cm]{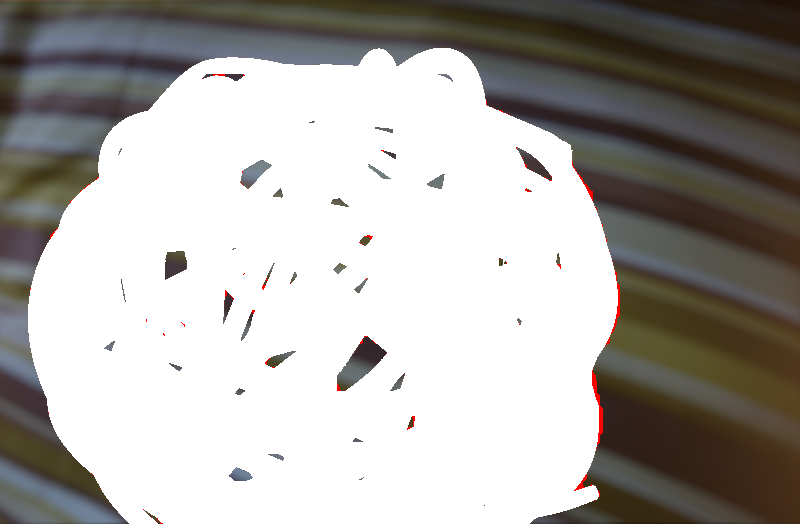}} &
\subfloat{\includegraphics[height=2cm]{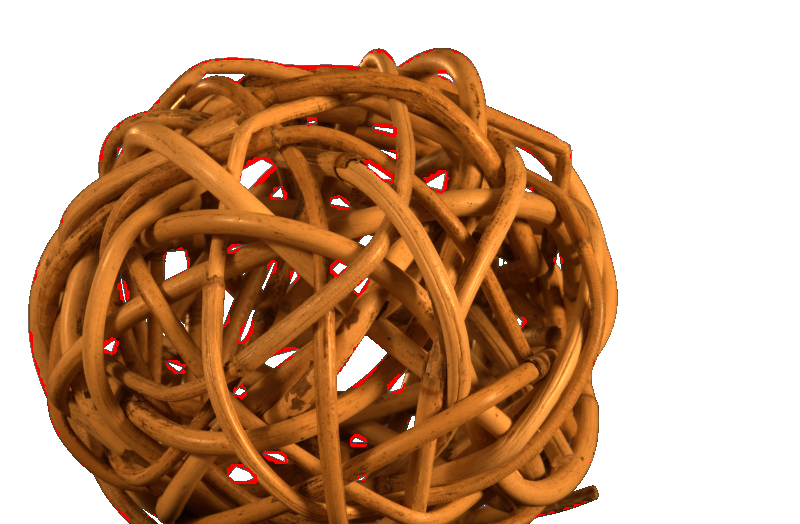}} \\
\multicolumn{2}{c}{Proposed Method - Error Rate: 0.52\%} &
\multicolumn{2}{c}{Reference Method - Error Rate: 1.27\%} \\
\\
\multicolumn{4}{c}{gt07} \\
\subfloat{\includegraphics[height=2cm]{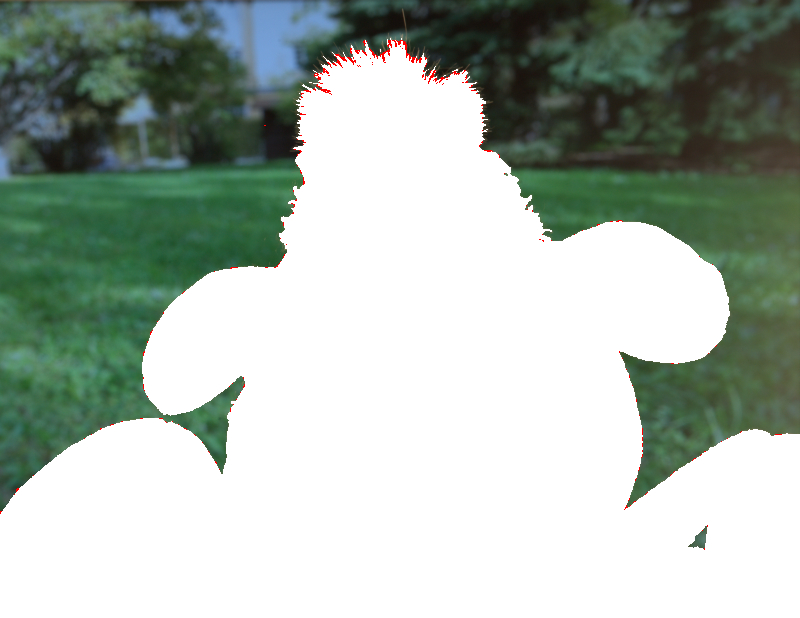}} &
\subfloat{\includegraphics[height=2cm]{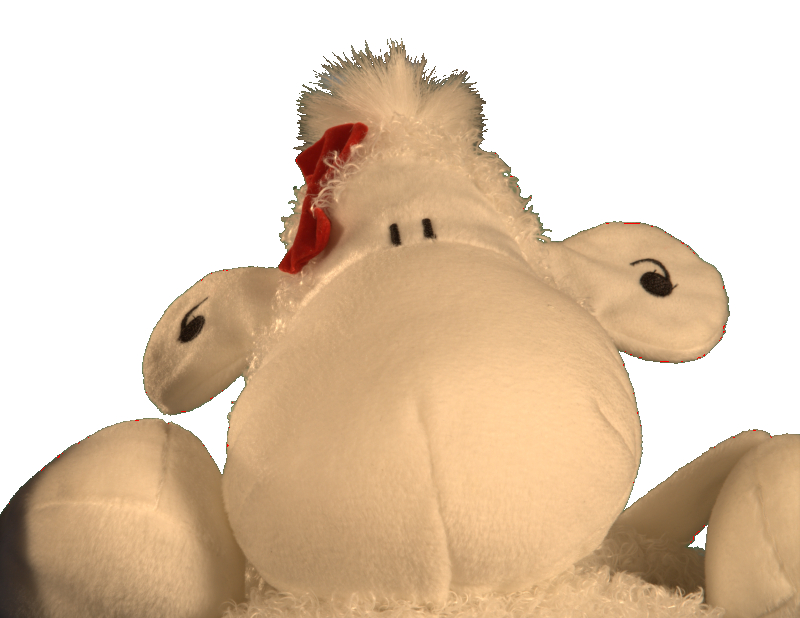}} &
\subfloat{\includegraphics[height=2cm]{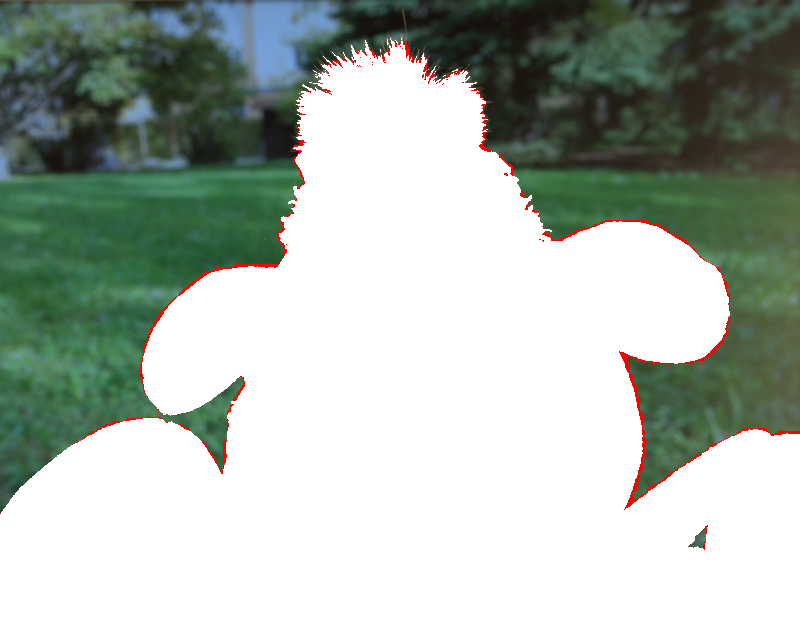}} &
\subfloat{\includegraphics[height=2cm]{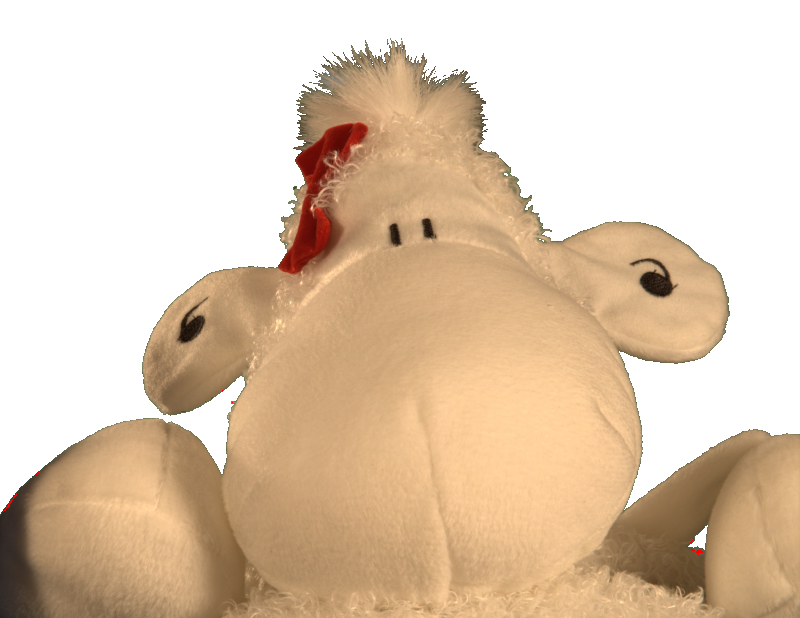}} \\
\multicolumn{2}{c}{Proposed Method - Error Rate: 0.21\%} &
\multicolumn{2}{c}{Reference Method - Error Rate: 0.64\%} \\
\\
\multicolumn{4}{c}{gt13} \\
\subfloat{\includegraphics[height=2cm]{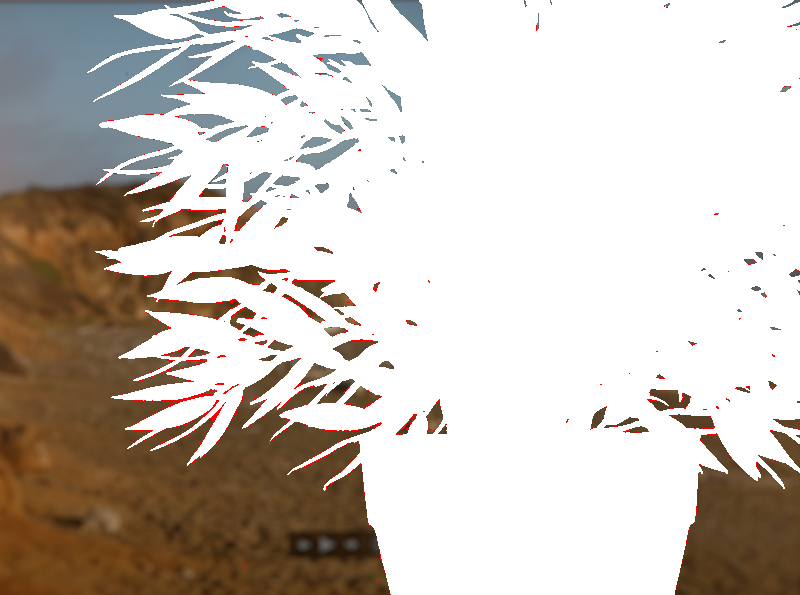}} &
\subfloat{\includegraphics[height=2cm]{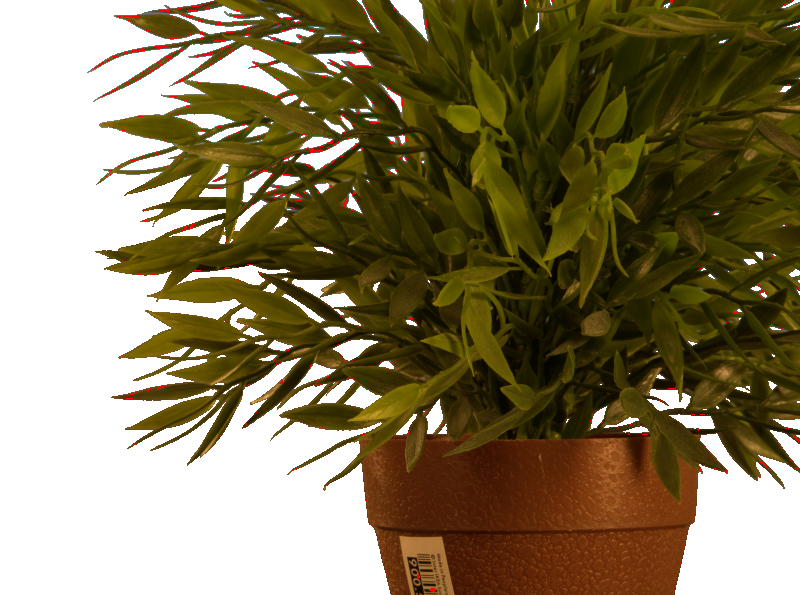}} &
\subfloat{\includegraphics[height=2cm]{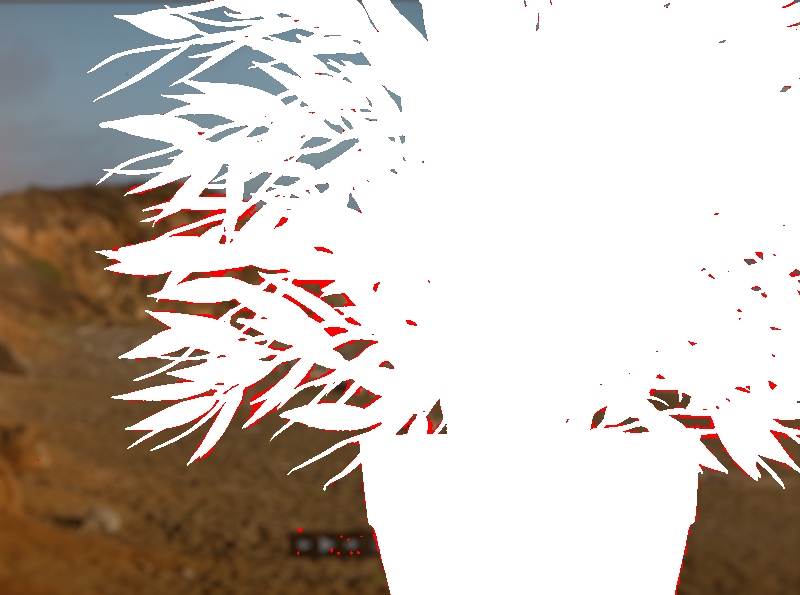}} &
\subfloat{\includegraphics[height=2cm]{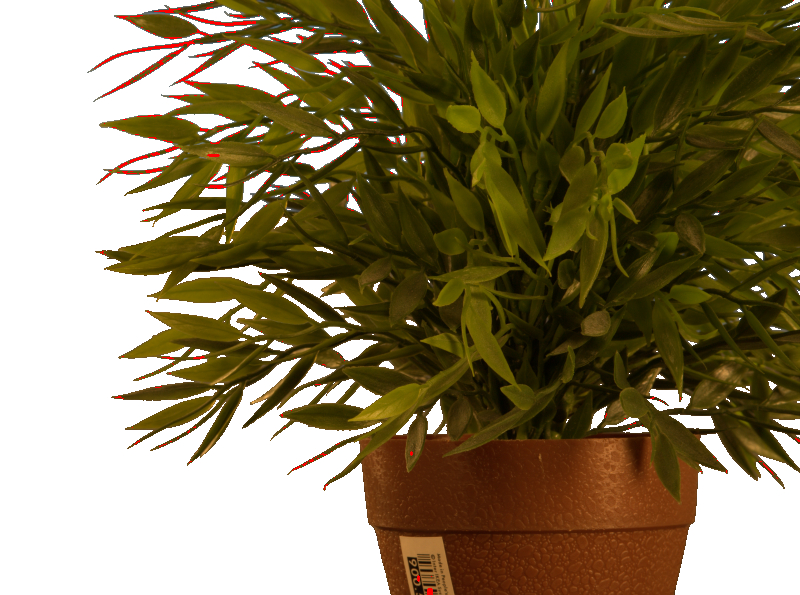}} \\
\multicolumn{2}{c}{Proposed Method - Error Rate: 1.08\%} &
\multicolumn{2}{c}{Reference Method - Error Rate: 2.11\%} \\
\\
\multicolumn{4}{c}{cross} \\
\subfloat{\includegraphics[height=2cm]{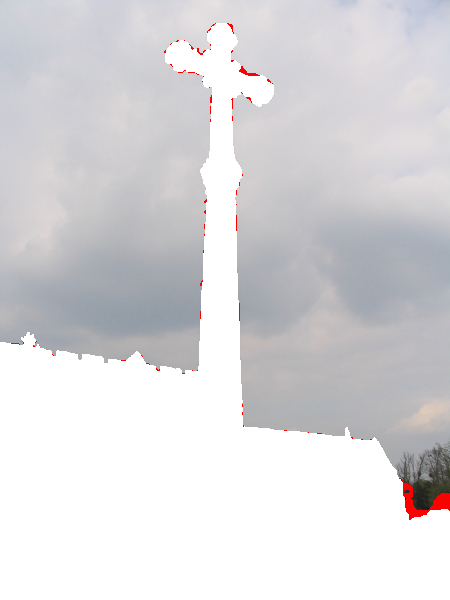}} &
\subfloat{\includegraphics[height=2cm]{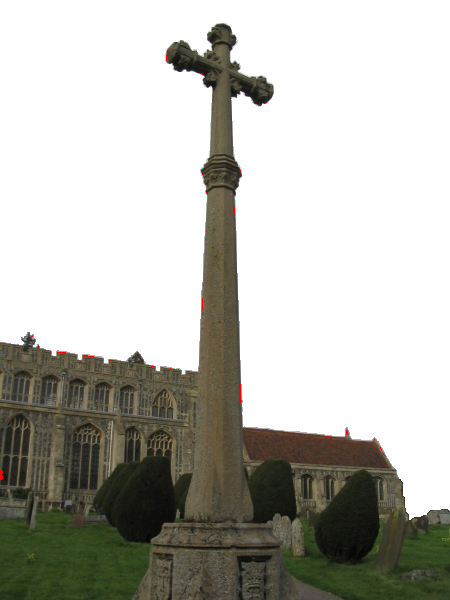}} &
\subfloat{\includegraphics[height=2cm]{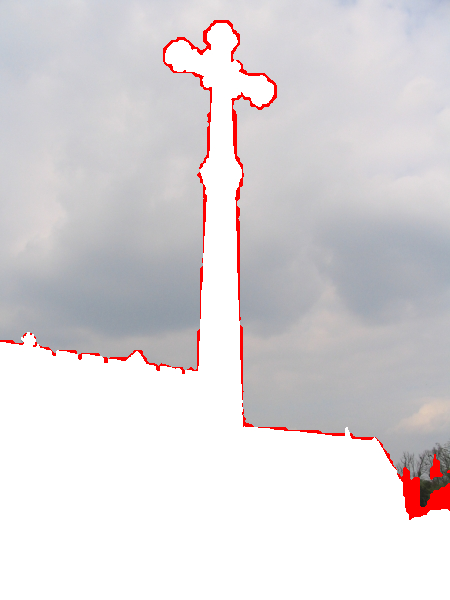}} &
\subfloat{\includegraphics[height=2cm]{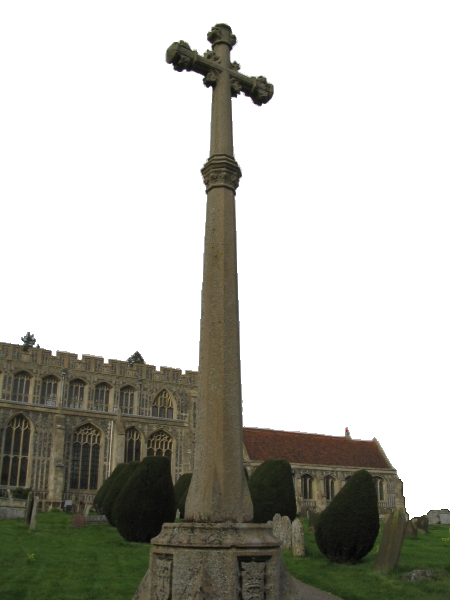}} \\
\multicolumn{2}{c}{Proposed Method - Error Rate: 0.48\%} &
\multicolumn{2}{c}{Reference Method - Error Rate: 1.79\%} \\

\end{tabular}
\caption{Image segmentation results by the proposed model and the reference model without the use of the cut polygon tool. Misclassified pixels are highlighted in red.}
\label{fig:ResultCutPolygon}
\end{figure}

The proposed method obtained the best pixel classification accuracy in comparison to the reference method. In the average, the proposed method obtained an error rate of 0.64\% while the reference method obtained an error rate of 1.72\%. The processed images can be observed in Figure \ref{fig:ResultCutPolygon}.

Regarding processing time, the reference method took 1371.08 seconds to output the results presented in Figure \ref{fig:ResultCutPolygon} while the proposed method took 1205.06 seconds.

Table \ref{tab:ResultBest} shows the five data set images in which the proposed method obtained the lowest error rates. In this case, all the images used the cut polygon feature during the proposed model execution to cut the segmentation area of interest, whilst in the reference method, the entire image was used. This cut process used in the images resulted in smaller networks for the proposed method in comparison to the reference method. The average of the images' complex network nodes, described in Table \ref{tab:ResultBest}, was of 2,089 nodes with 554,455 edges in the proposed method whilst, for the reference method, it was of 4,592 nodes and 2,102,821 edges.

\begin{table}[htb]
\centering
\caption{Error rates in the images with the lowest error rates achieved by the proposed method.}
\begin{tabular}{lrr}
\hline
Image name            & Proposed  & Reference \\
\hline
Monitor\_2007\_003011 & 0.02\%    & 1.09\%     \\
Train\_2007\_004627   & 0.09\%    & 0.76\%     \\
Car\_2008\_001716     & 0.10\%    & 2.51\%     \\
Monitor\_2007\_004193 & 0.11\%    & 3.00\%     \\
Person\_2007\_002639  & 0.12\%    & 2.47\%     \\ \hline
Average                 & 0.08\%    & 1.94\%     \\ \hline
\end{tabular}
\label{tab:ResultBest}
\end{table}

\begin{figure}[]
\centering
\setlength\tabcolsep{1.5pt}
\begin{tabular}{cccc}
\multicolumn{4}{c}{monitor_2007_003011} \\
\subfloat{\includegraphics[height=1.75cm]{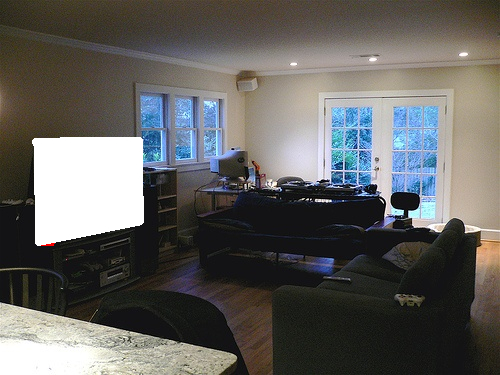}} &
\subfloat{\includegraphics[height=1.75cm]{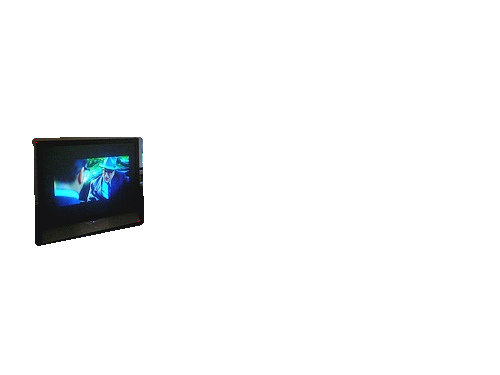}} &
\subfloat{\includegraphics[height=1.75cm]{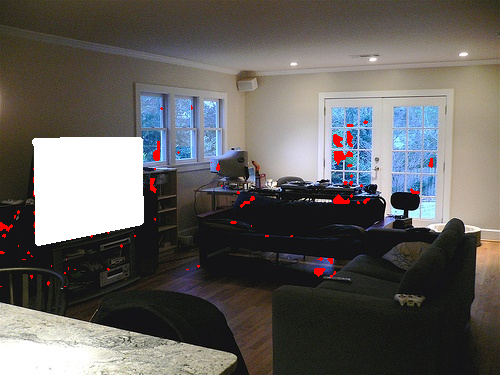}} &
\subfloat{\includegraphics[height=1.75cm]{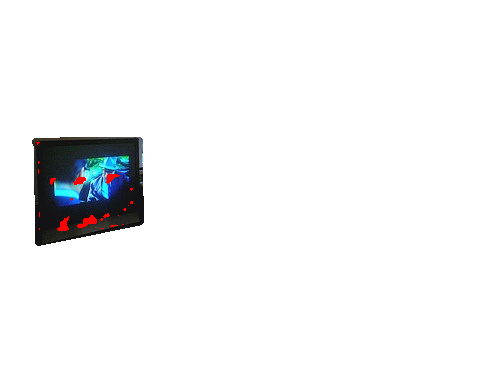}} \\
\multicolumn{2}{c}{Proposed Method - Error Rate: 0.02\%} &
\multicolumn{2}{c}{Reference Method - Error Rate: 1.09\%} \\
\\
\multicolumn{4}{c}{train_2007_004627} \\
\subfloat{\includegraphics[height=1.75cm]{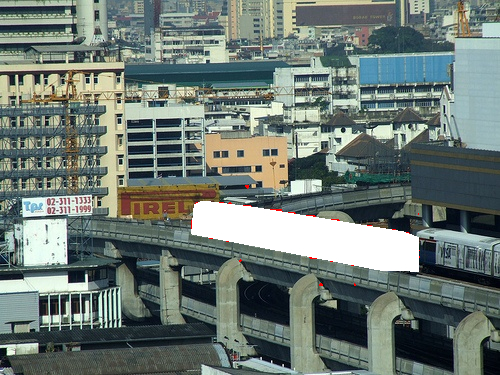}} &
\subfloat{\includegraphics[height=1.75cm]{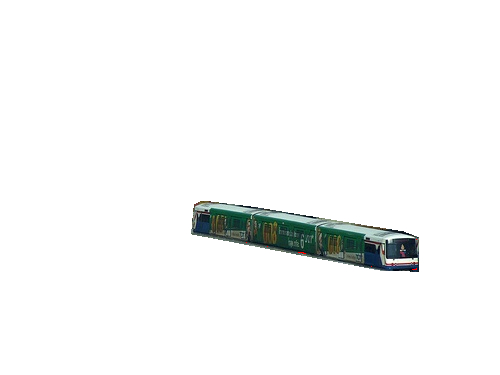}} &
\subfloat{\includegraphics[height=1.75cm]{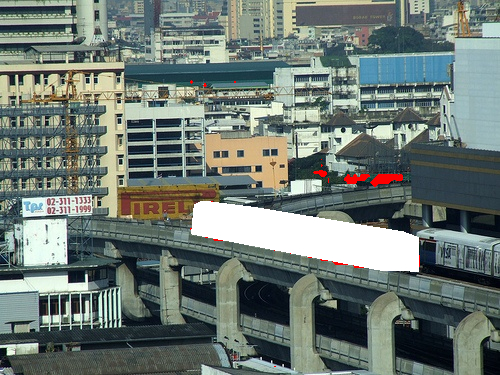}} &
\subfloat{\includegraphics[height=1.75cm]{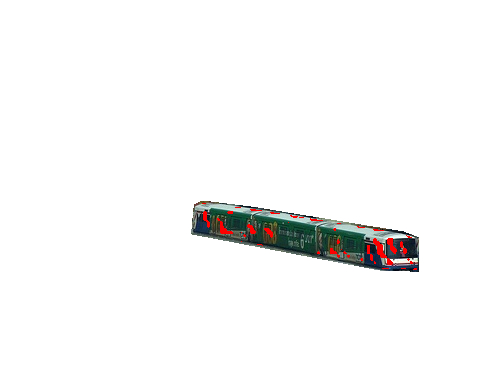}} \\
\multicolumn{2}{c}{Proposed Method - Error Rate: 0.09\%} &
\multicolumn{2}{c}{Reference Method - Error Rate: 0.76\%} \\
\\
\multicolumn{4}{c}{car_2008_001716} \\
\subfloat{\includegraphics[height=1.75cm]{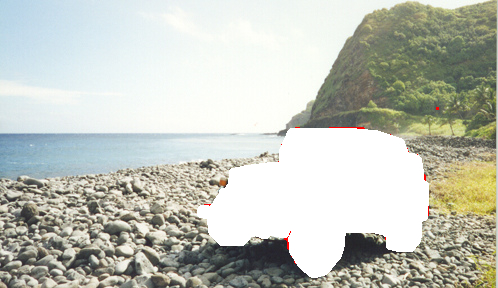}} &
\subfloat{\includegraphics[height=1.75cm]{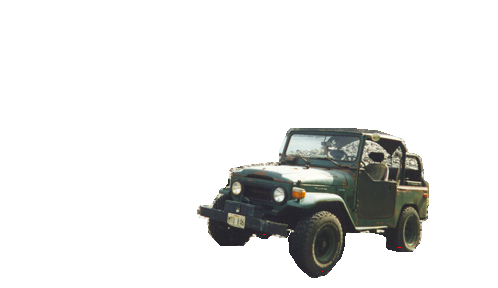}} &
\subfloat{\includegraphics[height=1.75cm]{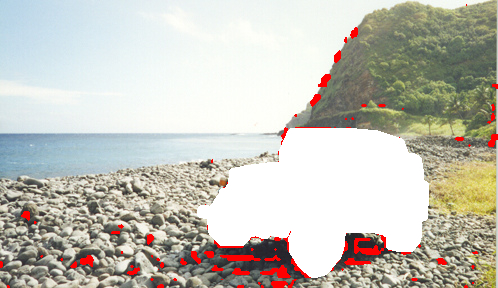}} &
\subfloat{\includegraphics[height=1.75cm]{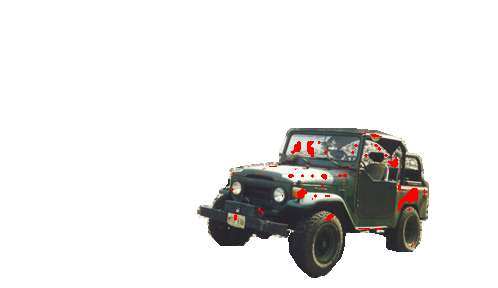}} \\
\multicolumn{2}{c}{Proposed Method - Error Rate: 0.10\%} &
\multicolumn{2}{c}{Reference Method - Error Rate: 2.51\%} \\
\\
\multicolumn{4}{c}{monitor_2007_004193} \\
\subfloat{\includegraphics[height=1.75cm]{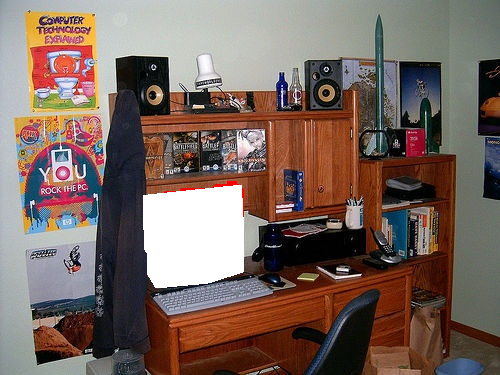}} &
\subfloat{\includegraphics[height=1.75cm]{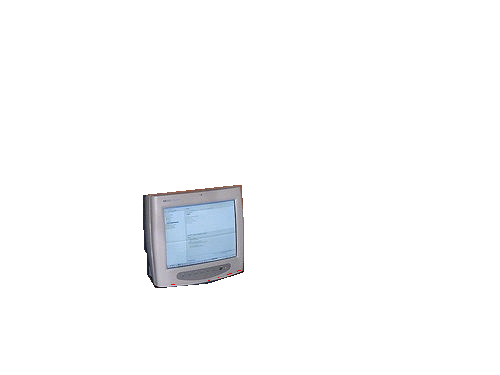}} &
\subfloat{\includegraphics[height=1.75cm]{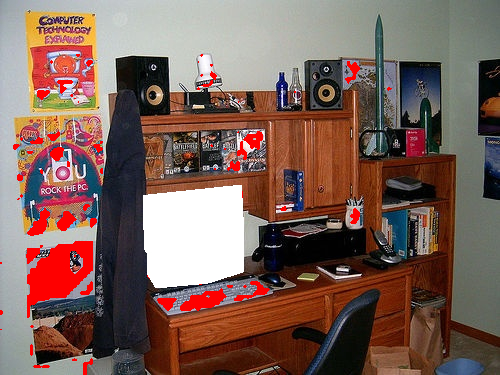}} &
\subfloat{\includegraphics[height=1.75cm]{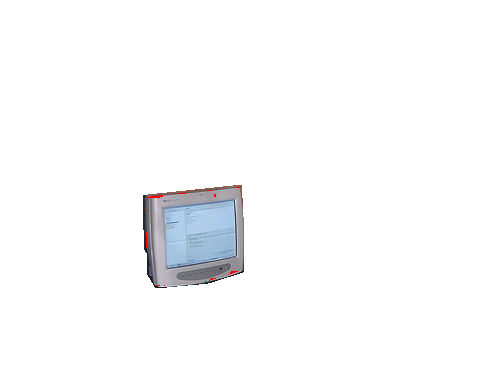}} \\
\multicolumn{2}{c}{Proposed Method - Error Rate: 0.11\%} &
\multicolumn{2}{c}{Reference Method - Error Rate: 3.00\%} \\
\\
\multicolumn{4}{c}{person_2007_002639} \\
\subfloat{\includegraphics[height=1.75cm]{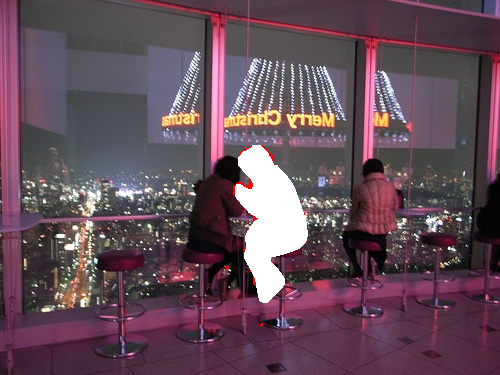}} &
\subfloat{\includegraphics[height=1.75cm]{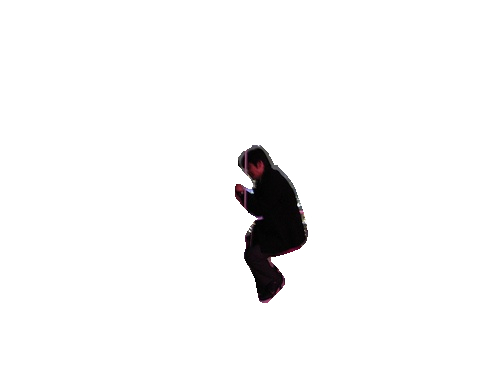}} &
\subfloat{\includegraphics[height=1.75cm]{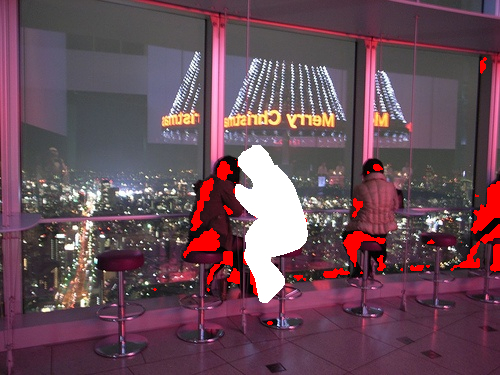}} &
\subfloat{\includegraphics[height=1.75cm]{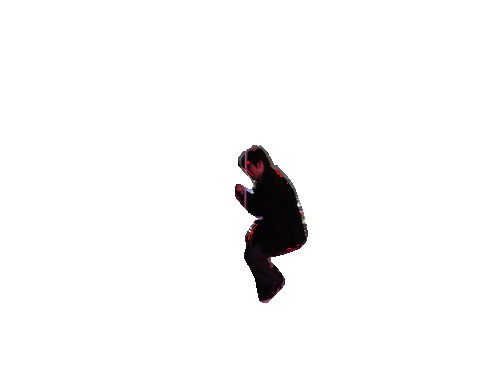}} \\
\multicolumn{2}{c}{Proposed Method - Error Rate: 0.12\%} &
\multicolumn{2}{c}{Reference Method - Error Rate: 2.47\%} \\

\end{tabular}
\caption{Images segmentation results by the proposed model with the use of the cut polygon tool and the reference model. Misclassified pixels are highlighted in red.}
\label{fig:ResultBest}
\end{figure}

By analyzing Table \ref{tab:ResultBest}, it can be verified that the proposed method, with a 0.08\% error rate, obtained a superior performance when compared to the reference method, which obtained an error rate of 1.94\%. Regarding the average execution time, the proposed method performed the images' segmentation in 255.46 seconds whilst the reference method performed it in 823.30 seconds. The higher execution time difference is justified by the difference in the complex network generation which defined smaller networks for the proposed method. These processed images can be observed in Figure \ref{fig:ResultBest}.

The general performance of both methods, considering the processing of all the $151$ images, is shown in Table \ref{tab:ResultBase} and Table \ref{tab:ResultErroBase}. In Table \ref{tab:ResultBase}, the size of the generated networks, for each method, can be verified. The proposed method generated smaller networks, in both nodes and edges, in comparison to the reference method. The $151$ images studied had an average of $200,124$ pixels, from which $2,783$ are unlabeled. The proposed method generated complex networks with an average of $7,538$ nodes and $838,564$ edges, whilst the reference method generated complex networks with a higher average of $17,946$ nodes and $2,354,555$ edges.

\begin{table}[htb]
\centering
\caption{Generated complex networks average characteristics by the proposed model and the reference model.}
\begin{tabular}{l|cc|ccc}
\hline
           & \multicolumn{2}{c}{\# Pixels} & \multicolumn{3}{c}{Characteristics} \\ \cline{2-6}
Method     & All       & Unlabeled & Particles  & Nodes      & Edges     \\ \hline
Proposed   & 200,124   & 2,783     & 2,860      & 7,538      & 838,564      \\
Reference  & 200,124   & 2,783     & 5,487      & 17,946     & 2,354,555     \\ \hline
\end{tabular}
\label{tab:ResultBase}
\end{table}

Table \ref{tab:ResultErroBase} shows the average error rate when classifying the pixels for all the $151$ images studied, using both the proposed method and the reference method. The proposed method achieved an average error rate of 0.49\% while the reference method achieved 3.14\%. Regarding execution time, the proposed method was beneficiated by the generation of smaller networks. It took an average of 432.54 seconds to segment the images, whilst the reference method performed the task in an average time of 1082.94 seconds.

\begin{table}[htb]
\centering
\caption{Average error rate and execution time in the $151$ images by the proposed method and the reference method.}
\begin{tabular}{lrr}
\hline
Method     & Error Rate & Time(s)   \\ \hline
Proposed   & 0.49\% & 432.54  \\
Reference & 3.14\% & 1082.94 \\ \hline
\end{tabular}
\label{tab:ResultErroBase}
\end{table}

\begin{figure}[htb]
\centering
\setlength\tabcolsep{1.5pt}
\begin{tabular}{c}
\subfloat{\includegraphics[height=8cm]{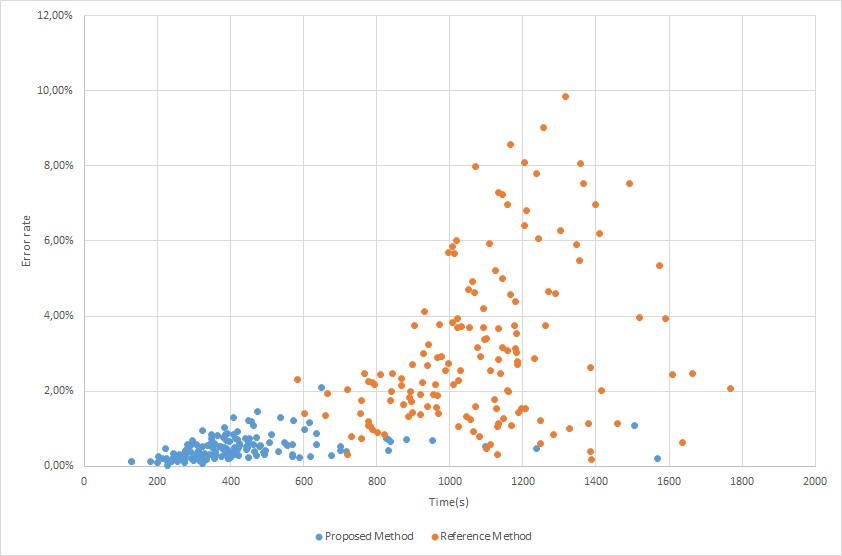}}
\end{tabular}
\caption{Relation Analysis between error rate and processing time presented by the proposed model and reference model.}
\label{fig:AnalyzeTimeErroRate}
\end{figure}

Figure \ref{fig:AnalyzeTimeErroRate} shows the relationship between the error rate and the execution time to perform the segmentation of each of the $151$ images. The proposed method has better stability in comparison to the reference method. The images processed by the proposed method are concentrated in a specific region of the graphic, whilst the reference method image's processing results are spread through the graphic. Therefore, the proposed method is less susceptible to features variations among the different images presented to it in this study.

\section{Conclusions}
\label{sec:Conclusions}

Autonomous segmentation represents one of the most complex image processing tasks since there is not an image segmentation algorithm that can be used in every situation. This paper presented a methodology to improve the automation level of the particle competition and cooperation model for image segmentation.

The proposed approach eliminates the weight vector $\lambda$ from the model since it is a parameter that needed to be set for each image to be segmented, requiring some expertise of the user. This elimination increased the model automation and it is achieved by optimizing the network construction phase, without changes in the particle competition and cooperation iterative phase.

The computer simulations have shown that the proposed method achieved better results when compared to the reference method, both in accuracy and performance. The proposed method obtained an average error rate of only 0.49\%, whilst the reference method obtained an average error rate of 3.14\% for the $151$ images studied.

The processing time was also reduced. The proposed method performed the segmentation in an average time of 432.54 seconds, whilst the reference method took an average time of 1082.94 seconds. This fact is justified by the average size of the networks generated by the studied models. The proposed method generated networks significantly smaller than the reference method.

\bibliographystyle{splncs04}
\bibliography{iccsa2020}

%
%
%
%

\end{document}